\documentclass[letterpaper, 10 pt, conference]{ieeeconf}  % Comment this line out if you need a4paper
\IEEEoverridecommandlockouts
\usepackage{cite}
\usepackage{amsmath,amssymb,amsfonts}
\usepackage{algorithm}
\usepackage[noend]{algpseudocode}
\usepackage{graphicx}
\usepackage{textcomp}
\usepackage{xcolor}
\usepackage{svg}
\usepackage[nolist]{acronym}
\usepackage{subcaption}
\usepackage{units}
\usepackage{comment}
\usepackage{hyperref}
\usepackage{gensymb}
\usepackage[nolist]{acronym}
\usepackage{booktabs}
\usepackage{lipsum}
\usepackage{mathtools}
\usepackage{microtype}

\definecolor{todo-red}{RGB}{200,12,12}
\definecolor{green4}{RGB}{0,128,0}

\newcommand{\etal}{\textit{et al}.}
\newcommand{\ie}{\textit{i.e.}}
\newcommand{\eg}{\textit{e.g.}}

\newcommand{\segmap}{\textit{SegMap}}
\newcommand{\semsegmap}{\textit{SemSegMap}}
\renewcommand{\vec}[1]{\mathbf{#1}}

\begin{document}
\title{\semsegmap{} -- 3D Segment-based Semantic Localization}

\author{Andrei Cramariuc$^{1,\ast}$, Florian Tschopp$^{1,\ast}$, Nikhilesh Alatur$^1$, Stefan Benz$^2$, Tillmann Falck$^2$,\\ Marius Brühlmeier$^{1}$, Benjamin Hahn$^1$, Juan Nieto$^3$, and Roland Siegwart$^1$%
\thanks{$^\ast$Authors contributed equally to this work}%
\thanks{$^1$Authors are members of the Autonomous Systems Lab, ETH Zurich, Switzerland; {\tt\small \{firstname.lastname\}@mavt.ethz.ch}}%
\thanks{$^2$Authors are members of Robert Bosch GmbH, Germany; {\tt\small \{firstname.lastname\}@de.bosch.com}}%
\thanks{$^3$Author is with Microsoft, Switzerland but the work was done while the author was a member of $^1$.}%
\thanks{This work was supported by Robert Bosch GmbH, Germany. The code is available at \url{https://github.com/ethz-asl/segmap}.} %
}

\maketitle
\thispagestyle{empty}
\pagestyle{empty}

\begin{abstract}
Localization is an essential task for mobile autonomous robotic systems that want to use pre-existing maps or create new ones in the context of \acs{slam}.
Today, many robotic platforms are equipped with high-accuracy 3D \acs{lidar} sensors, which allow a geometric mapping, and cameras able to provide semantic cues of the environment.
Segment-based mapping and localization have been applied with great success to 3D point-cloud data, while semantic understanding has been shown to improve localization performance in vision based systems.
In this paper we combine both modalities in \semsegmap, extending \segmap{} into a segment based mapping framework able to also leverage color and semantic data from the environment to improve localization accuracy and robustness.
In particular, we present new segmentation and descriptor extraction processes.
The segmentation process benefits from additional distance information from color and semantic class consistency resulting in more repeatable segments and more overlap after re-visiting a place.
For the descriptor, a tight fusion approach in a deep-learned descriptor extraction network is performed leading to a higher descriptiveness for landmark matching.
We demonstrate the advantages of this fusion on multiple simulated and real-world datasets and compare its performance to various baselines.
We show that we are able to find $\unit[50.9]{\%}$ more high-accuracy prior-less global localizations compared to \segmap{} on challenging datasets using very compact maps while also providing accurate full 6~\acs{dof} pose estimates in real-time.
\end{abstract}

\section{Introduction}
\label{sec:introduction}
Mobile robots are continuously increasing their impact on our everyday life and becoming more and more viable not only in structured factories but also unstructured environments and in contact with humans~\cite{Cadena2016}. 
One of the most crucial capabilities of mobile robots is the ability to know their position in the environment in order to navigate and fulfill their task.
Positioning can be framed either in the context of localization in a known map or in the context of the \ac{slam} problem where localizations and potential loop closures are needed to maintain a consistent map.
A multitude of solutions to the positioning problem exist, mainly depending on the available sensor data, specific challenges of the environment and computational limitations of the robotic platform~\cite{Cadena2016}.

For standard indoor applications, visual localization based on hand-crafted keypoint descriptors~\cite{Schneider2017, Mur-Artal2015} has been demonstrated to achieve high accuracy and recall.
However, outdoor applications typically pose challenges to those methods, namely large-scale environments, self similarity and vast appearance changes due to weather, daytime and seasonal conditions~\cite{Milford2012}.
Vision based learning methods improve on viewpoint and appearance invariance~\cite{RevaudR2D2:Descriptor, Sarlin2018FromScale, Arandjelovic2015} by enabling a more context aware description. 
In contrast, \ac{lidar} based localization achieves illumination invariance using geometry to describe the environment~\cite{dube2019segmap, Elhousni2020AVehicles}, however in turn missing rich information available from vision.

\begin{figure}
    \centering
    \includegraphics[width=\columnwidth]{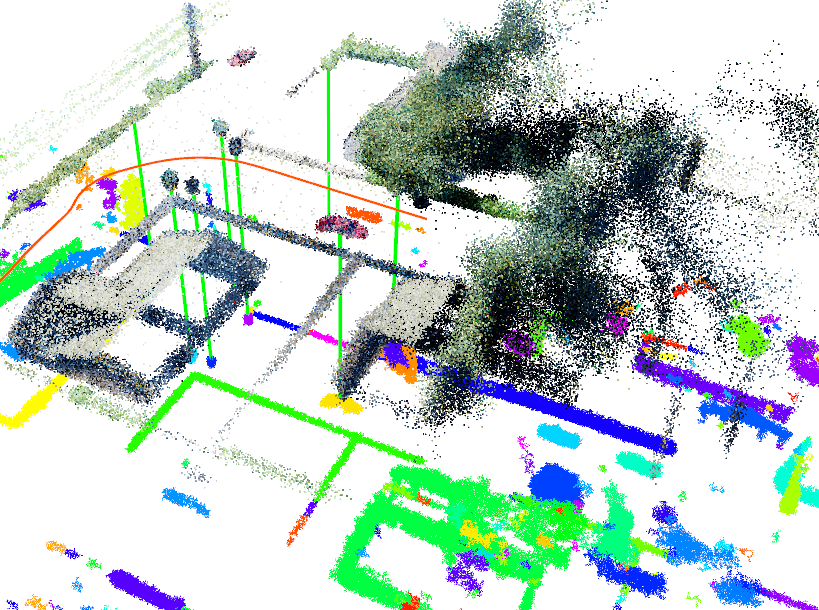}
    \caption{This image shows the \semsegmap{} pipeline in action. \semsegmap{} is able to perform segment-based semantic localization on \acl{pc} data enriched with semantic and color information from a visual camera. The currently observed local map around the robot is shown as the colored \acl{pc} on top of the global map depicted below, with each segment having a unique color. Green lines show matched segment correspondences leading to a localization while the orange line shows the robot trajectory.}
    \label{fig:teaser}
    \vspace{-0.3cm}
\end{figure}

As a combined solution, in this paper, we introduce \semsegmap, a method that leverages the visual and semantic information available from cameras and fuses it with geometric information from a standard 3D \ac{lidar}.
As a basis for our localization framework we use \segmap~\cite{dube2019segmap}, a \ac{lidar} based \ac{slam} pipeline that uses 3D segments of the environment as landmarks and allows for 6D pose retrieval from compact descriptors in large-scale maps.
In contrast to \segmap, in \semsegmap, as outlined in Figure~\ref{fig:teaser}, first the \ac{pc} gets enriched with color and semantic information using back-projection of semantically segmented RGB images. 
Further, the \ac{pc} is segmented based on geometric, color and semantic information to create consistent and meaningful segments.
We show in multiple experiments that as a result of this fusion, the segmentation process and the generated descriptors become more robust to viewpoint and appearance changes, thus enabling a more consistent re-localization of the robot.

Our contributions are as follow
\begin{itemize}
    \item We show that integrating color and semantic information from a cameras into \acp{pc} improves both the segmentation and descriptor generation processes, leading to more consistent 6D localizations in a SLAM pipeline.
    \item We introduce a simulation based learning pipeline for training segment descriptors using ground truth associations, and show their transferability to real-world scenarios.
    \item We demonstrate the performance of \semsegmap{} in an extensive evaluation on simulation and real-world data outperforming various baselines.
    \item For the benefit of the community, we open-source the whole framework under a permissive license available at: \url{https://github.com/ethz-asl/segmap}.
\end{itemize}

% We could remove that whole part if we need space.
%The remainder of the paper is organized as follows: In Section~\ref{sec:rel_work}, a summary of related work on visual, semantic and \ac{lidar}-based localization is presented. Section~\ref{sec:method} presents our data fusion framework, segmentation approach and descriptor retrieval. In Section~\ref{sec:experiments}, \semsegmap{} is evaluated on both simulated and real-world public datasets and compared to state-of-the-art baselines for \ac{pc} based localization. Finally, in Section~\ref{sec:conclusions}, we provide a conclusion of the presented work and a short outlook on future work.

\section{Related Work}
\label{sec:rel_work}
The ability to localize is at the heart of the SLAM problem~\cite{Cadena2016}.
Two distinct problems addressed in localization are the localization of landmarks, which can then be used to calculate a precise 6D location in a global map, and place recognition, where only a rough neighborhood is estimated.

Vision-based place recognition methods such as NetVLAD~\cite{Arandjelovic2015} or DELF~\cite{noh_large-scale_2017} have the advantage that they can incorporate a lot of contextual information and thus gain high robustness to viewpoint and illumination changes.
Some recent techniques explicitly model the semantics of the scene to obtain higher robustness towards seasonal changes~\cite{Gawel2017,benbihi_image-based_2020,hu2020dasgil}. 
More precise visual keypoint-based place recognition methods are well studied~\cite{lowry_visual_2016} and of significant interest~\cite{noauthor_benchmarking_nodate}, but they present their own set of challenges with regards to scalability, viewpoint, and illumination changes.
A compromise between precise keypoint localization and the ability to incorporate contextual and semantic information can be found in object-based localization frameworks \cite{Ok2019RobustNavigation,taubner_lcd_2020}.

\ac{lidar} based localization methods rely mainly on matching geometry and can be separated into various categories.
Many \ac{pc} registration methods, out of which \ac{icp}~\cite{besl1992method} is the most well-known one, require a good pose prior and are not suited for global localization.
While global registration methods exist that work beyond the local context~\cite{bernreiter2021phaser, Le_2019_CVPR}, they still require storing at least parts of the \ac{pc} data.
This can be partially mitigated by only extracting compact descriptors during map building and localization.
Descriptors of single \ac{lidar} scans~\cite{chen_overlapnet_2020, yin20193d, kim_1-day_2019, Yin2018LocNet} only allow rough location estimates and not precise 6D poses that could be integrated into a \ac{slam} pipeline.
Descriptors of local keypoint neighborhoods~\cite{Rusu2009FPFH,lu2019l3,kallasi2016fast,gojcic2019perfect} in turn can be noisy and lack distinctiveness due to only having geometry data available.
A combination can be achieved by performing a refinement step, \eg{} using \ac{icp}, after the rough localization, but would again require storing and working with \ac{pc} data~\cite{schaupp_oreos_2019,zaganidis_semantically_2019}.

\ac{pc} segment description and mapping based approaches were first proposed by Douillard \etal~\cite{douillard2012scan} and Nieto \etal~\cite{nieto2006scan} and then further extended into a full \ac{slam} pipeline in \segmap{} by Dub{\'e} \etal~\cite{dube2019segmap}.
\segmap{} leverages advantages from both local and global descriptors by looking at features in the environment that are large enough to be more robust and meaningful, while also maintaining the ability to produce accurate 6D poses.
Extensions to the pipeline include different training methods for the descriptor, as well as fine tuning to different environments~\cite{tinchev_seeing_2018,tinchev_learning_2019}.

% Combined Lidar and Camera (semantics)
Enriching \acp{pc} with semantic information, \eg{} obtained by fusing camera and \ac{lidar} data, has proven beneficial~\cite{sun_recurrent-octomap_2018,zaganidis_integrating_2018,chen_suma_2019, parkison_semantic_2018, Bernreiter2021S2Loc}.
Ratz~\etal~\cite{Ratz2020Oneshot} propose not only to use just one scan instead of an accumulated \ac{pc} but also to enrich the descriptor with appearance information from a camera using NetVLAD and a customized fusion layer in their network.
This increases accuracy at the cost of a significant decrease in computational performance, due to the expensive NetVLAD backbone.
Also, the segmentation procedure is still purely based on geometry and cannot incorporate any additional information from the camera.
Sch\"onberger~\etal~\cite{Schonberger2017} propose a localization scheme based on semantically labeled 3D \acp{pc} able to localize in the presence of severe viewpoint and appearance changes.
They extract a deep-learned descriptor for a subvolume of the semantic map and compare it to similarly extracted ones from a prior map.
However, using such submaps only provides a rough place-recognition like localization and require further refinement steps to get 6D poses.
Furthermore, by operating on comparably larger subvolumes and the need to check multiple hypothesis, the corresponding descriptor network is computationally expensive. 

In contrast, \semsegmap{} introduces a framework that is both efficient enough to be run in real-time on a consumer CPU (excluding the semantic segmentation running on the GPU) and still able to leverage the rich information available from geometry, appearance and semantics. 
We perform a very early fusion of the two sensor modalities which allows also the segmentation to be based on all available information.

\section{\semsegmap}
\label{sec:method}
\begin{figure*}[ht]
    \centering
    \includegraphics[width=1.8\columnwidth]{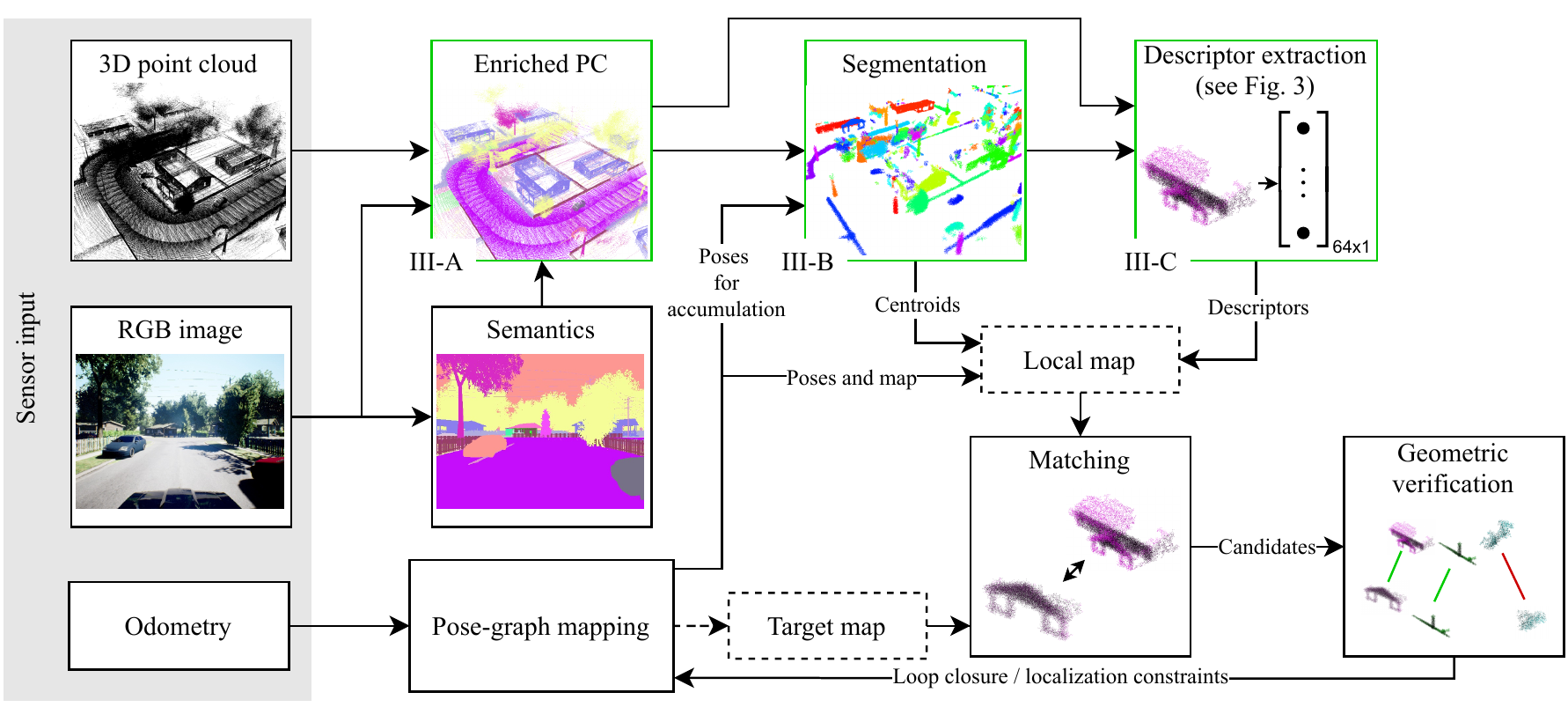}
    \caption{Overview of the \semsegmap{} pipeline. The whole pipeline can be run in \textit{localization mode} with a target map loaded from the disk or in \textit{loop closure mode} where the target map is provided by the current pose-graph. Green: Main modules changed from~\cite{dube2019segmap} with corresponding section numbers.}
    \label{fig:pipeline}
    \vspace{-0.3cm}
\end{figure*}

In this section we present the details of the proposed \semsegmap{} pipeline, as shown in Figure~\ref{fig:pipeline}. 
The approach can be split into several key modules, out of which the new segmentation and descriptor explained in Sections~\ref{sec:semseg} and~\ref{sec:method-description}, represent the core contributions of this paper.

\subsection{Semantic enrichment}
\label{sec:method-segmentation}
The inputs to the pipeline consist of a stream of color images and \acp{pc}.
The color images are passed through a semantic segmentation network to obtain a semantic class for each pixel.
Using the extrinsic calibration between the camera and the \ac{lidar} as well as the intrinsic calibration parameters of the camera, the color and semantic class of each pixel is projected onto the \ac{pc}.
The result is a set of enriched points in the form $\vec{p} = \{x, y, z, h, s, v, c\}$, where $x$, $y$, and $z$ are the spatial coordinates, featuring also color values $h$, $s$, and $v$ in HSV space (result visible in Figure~\ref{fig:teaser}), and semantic class labels $c$ (example depicted in Figure~\ref{fig:pipeline}: Enriched \ac{pc}).

\subsection{Semantic segmentation}
\label{sec:semseg}
To remove noise and achieve a more compact data representation, we accumulate the enriched \ac{pc} data in a fixed size voxel grid.
The voxel grid is a cylinder with a radius of $R$ that dynamically follows the robot and is centered on it.
For each voxel, the color information of multiple points is fused by using a running average over the incoming values to obtain the current value for the voxel.
In contrast, the semantic class labels can not be averaged, and therefore, all values are stored and the semantic label of the voxel is determined by majority voting.
Further filtering can be done by excluding points that belong to known dynamic classes \eg{} humans and cars. 

We use an incremental Euclidean segmentation that does not need to be rerun on the entire \ac{pc} at each step, but is computed incrementally only on the newly active voxels, as detailed in~\cite{dube2019segmap}.
A segment $\mathcal{S}$ is defined as a set of points, where for each point $\vec{p}_1 \in \mathcal{S}$ there exists at least one other point $\vec{p}_2 \in \mathcal{S}$, so that the distance between these two points is smaller than the segmentation distance $d_{segment}$.

To include the semantic and color information into the segmentation process, we modify the standard Euclidean distance function between two points $\vec{p}_1$ and $\vec{p}_2$ to be
\begin{equation}
\begin{split}
    d^2 =&\ (x_1-x_2)^2 + (y_1-y_2)^2 + (z_1-z_2)^2 + \\
    & f_h^2(|h_1-h_2|) + f_c^2(c_1, c_2).
\end{split}
\end{equation}
The function $f_h$ is used to compute the distance between two colors by only comparing the difference in hue values in order to mitigate the appearance variance and is defined as
\begin{equation}
    f_h(\Delta h) = \begin{cases*}
      p_h & $\min{(\Delta h, 1 - \Delta h)} > t_h $\\
      0 &\text{otherwise}\end{cases*},
\end{equation}
where $p_h$ is a fixed penalty for when the color difference is above a certain chosen threshold $t_h$ and where the hue values $h$ are normalized to $[0, 1)$.
In practice, there are two cases because the hue color space is cyclical.
Similarly, the semantic class distance function is defined as
\begin{equation}
    f_c(c_1, c_2) = \begin{cases*}
      p_c &\text{$c_1 \neq c_2$}\\
      0 &\text{otherwise}\end{cases*},
\end{equation}
where $p_c$ is a fixed penalty applied when the semantic classes do not match.
Fixed penalties are necessary because the color space and semantic spaces are either not continuous or not numerically comparable to physical distances in space.
A soft constraint on the segmentation can be imposed by choosing $p_h$ and $p_c$ smaller than $d_{segment}$, which means that two points that are sufficiently close in space can still be part of the same segment, even if they have a very different color or class.

During the segmentation process, at each step, the robot extracts a set of segments in the local map around itself.
Those segments slowly accumulate points as more observations are made from different viewpoints.
Similarly to how a keypoint is tracked, a segment will have multiple accumulated observations.
Therefore, the final observation, just before it moves out of the local map neighborhood, will be the most complete one.

\subsection{Description}
\label{sec:method-description}
For each segment observation, a learned descriptor is calculated and the local map is built by associating each descriptor to the corresponding segment centroid point.
For efficiency reasons, we only keep the descriptor of the last and most complete observation of each segment to create the target map for subsequent localizations or loop closures.

\begin{figure}[t]
    \centering
    \vspace{0.1cm}
    \includegraphics[width=0.9\columnwidth]{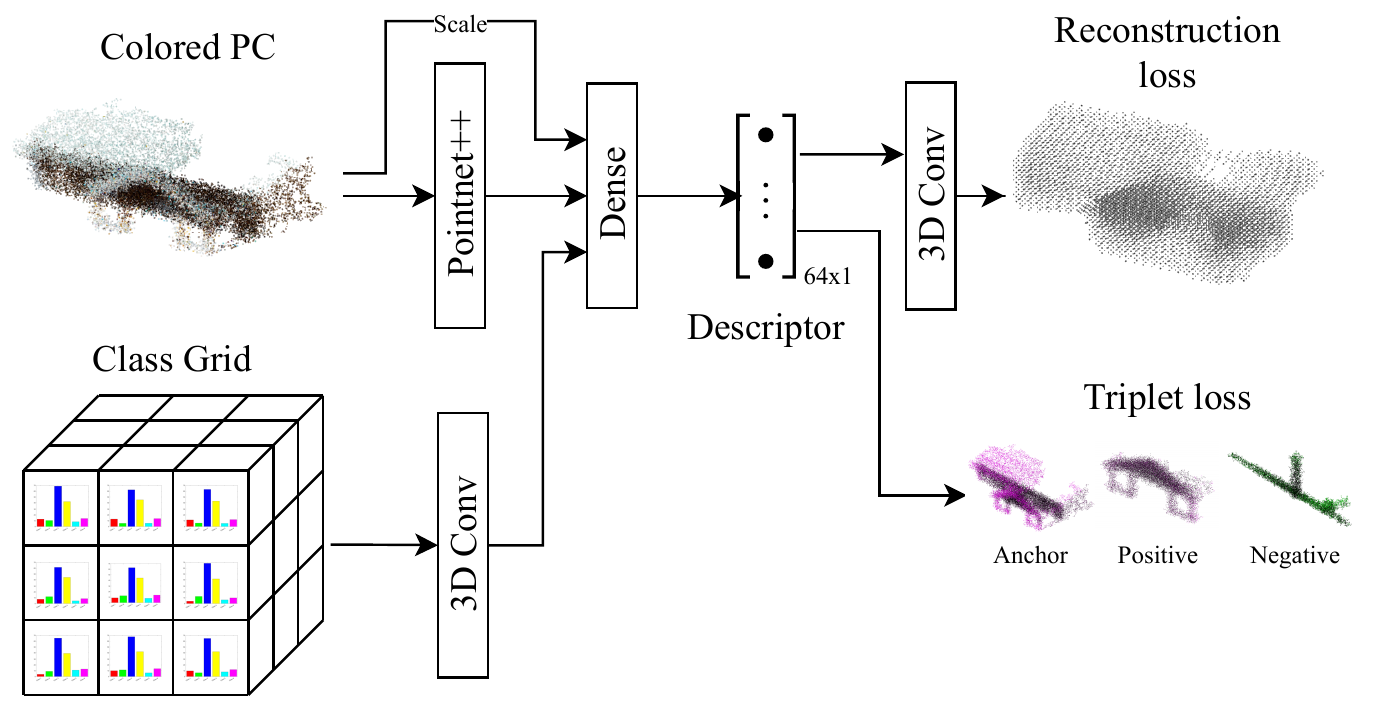}
    \caption{The descriptor extraction of the colored \ac{pc} is based on a Pointnet++~\cite{Qi2017Pointnetpp} backbone while the semantics are fused in a coarse voxel grid of semantic class histograms. The network is trained using both a reconstruction loss and a triplet loss.}
    \label{fig:descriptor_architecture}
        \vspace{-0.3cm}
\end{figure}

The new descriptor network, illustrated in Figure~\ref{fig:descriptor_architecture}, uses the Pointnet++ architecture~\cite{Qi2017Pointnetpp} that is based on hierarchical point set feature learning.
The input to the Pointnet++ backbone is the colored \ac{pc} segment that has been randomly subsampled to a fixed size of $2048$ points.
The class labels are instead accumulated into a very coarse $3\times 3\times 3$ spatial voxel grid, where each cell contains a normalized histogram of the class labels that fall inside that section of the \ac{pc} segment.
This very coarse description is enough because due to the semantic segmentation process the class labels inside most segments are relatively homogeneous.
For the sake of computational efficiency this class representation is handled separately by a small 3D convolutional network, whose output is later concatenated with that of the Pointnet++ backbone.
Finally, we also input into the network the scaling factor by which the point coordinates were normalized, which helps the network better discriminate between segments that do not match but are either visually or geometrically similar.

The descriptor is trained using both a triplet loss as well as a reconstruction loss from a convolutional decoder.
The triplet loss is formulated as
\begin{equation} \label{eq:triplet}
    L_\text{triplet} = \max(m + \sigma(\mathcal{A},\mathcal{P}) \cdot D_{\mathcal{A},\mathcal{P}} - D_{\mathcal{A},\mathcal{N}}, 0)
\end{equation}
where $m=0.4$ is the margin, $D_{\mathcal{A},\mathcal{P}}$ is the Euclidean distance between an anchor segment $\mathcal{A}$ and a positive example $\mathcal{P}$, $D_{\mathcal{A},\mathcal{N}}$ is the Euclidean distance between the anchor and a negative example $\mathcal{N}$, and $\sigma(\mathcal{A},\mathcal{P}) = \frac{\mathbf{card}(\mathcal{P})}{\mathbf{card}(\mathcal{A})}$ is the ratio in number of points between anchor and positive example.
%
%The $\sigma$ term is the size ratio in number of points between the anchor and positive example.
%
The scaling $\sigma(\mathcal{A},\mathcal{P})$ is a heuristic that prevents the loss function from penalizing too much segment observations that should match, but due to segment incompleteness only share a small overlap and therefore in practice are hard to match.
For the reconstruction loss we use a binary cross entropy loss applied to each voxel.
This enables approximate reconstructions of the \ac{pc} map from only the descriptor space for visualization and improves the descriptor quality of very similar looking segments.
During training we augment the \acp{pc} in multiple ways, including both geometric variations such as random rotations, jitter, scale shifts, missing points or sections, and visual variations such as color shifts or erronous class labels.

\subsection{Localization and loop closure}
\label{sec:localization}
To perform localization or loop closure, candidate correspondences are identified between the locally built map of segments and the prior or global map, respectively.
The candidates are identified using the descriptors of each locally visible segment and retrieving the $k$ most similar descriptors from the global map.
Finally, a geometric verification step is performed based on the centroids of the target and query match candidates to identify the 6~\ac{dof} transformation that leads to the largest set of inliers using \ac{ransac}.
%
% If the set of geometrically consistent matches is large enough, a 6~\ac{dof} transformation between the local and global map is calculated.
%
The resulting transformation can either be used as a localization result when localizing from a previously built map, or as a loop closure constraint in a \ac{slam} scenario.
In the latter case, both the loop closure constraint and robot odometry constraints are placed into an online pose graph based on iSAM2~\cite{Kaess2012iSAM2}.

\section{Experiments} \label{sec:experiments}
In this section, we describe our training procedure and present thorough evaluation of \semsegmap{} on both simulated and public real-world datasets, demonstrating a superior performance compared to different baselines on segmentation, descriptor quality, and localization accuracy and robustness. 
\subsection{Datasets}
Datasets including visual as well as \ac{pc} data, spanning large areas and covering different environmental conditions that include semantic annotations are rare and hard to obtain.
Therefore, for the data intensive step of descriptor training, we utilize simulated data to be able to quickly produce training data for \semsegmap.
Furthermore, simulated datasets provide the opportunity to specifically evaluate our contribution in isolation of sensor noise, state estimation and calibration inaccuracies, and imperfect semantic segmentation.
In addition, we demonstrate the transferablility of our approach to a challenging real-world dataset.
\subsubsection{Simulation}
Photo-realistic simulation is a popular tool for efficiently generating visually rich data with high quality ground truth annotations.
Some of the most popular simulation tools in the robotics domain are Gazebo \cite{Koenig2004gazebo}, CARLA \cite{Dosovitskiy17carla}, LGVSL \cite{Rong2020lgsvl} and AirSim \cite{Shah2017airsim}.
While CARLA and LGVSL focus on autonomous driving scenarios, Gazebo does not provide photorealistic visual output.
The datasets used in the following experiments were generated by AirSim using the "Modular Neighborhood Pack"\footnote{\url{https://www.unrealengine.com/marketplace/en-US/product/modular-neighborhood-pack}}, a large residential environment. 
Each dataset consists of RGB image data, a semantic segmentation map for the image, \ac{lidar} \acp{pc} as well as odometry information. 
The image data consists of three cameras, one on each side and a front facing one, spanning a total horizontal \ac{fov} of $\unit[270]{^\circ}$.
The simulated \ac{lidar} has a resolution of $1920\times64$ and the same $\unit[270]{^\circ}$ \ac{fov}.
All sensors are synchronized and operated at $\unit[5]{Hz}$.
\begin{figure}
    \centering
    \vspace{0.3cm}
    \includegraphics[width=0.75\columnwidth]{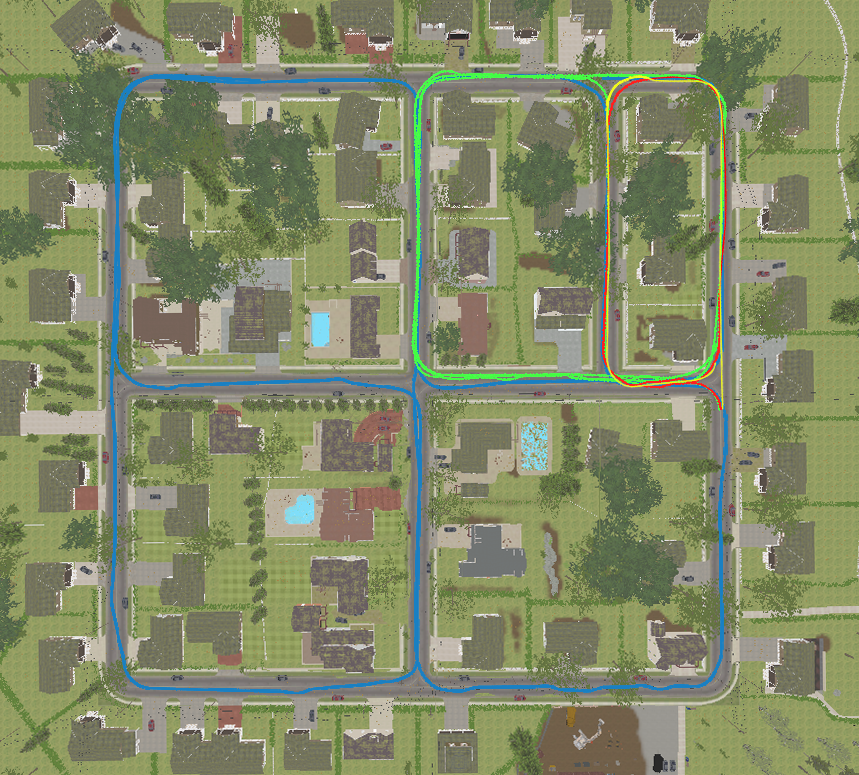}
    \caption{Bird's eye view of the simulation environment including the simulated trajectories (blue: \texttt{S0}, green: \texttt{S1}, red: \texttt{S2}, yellow: \texttt{S3}).}
    \label{fig:neighborhood-overview}
\end{figure}
\begin{figure}
    \centering
    \begin{subfigure}{.44\columnwidth}
        \centering
        \includegraphics[width=.95\columnwidth]{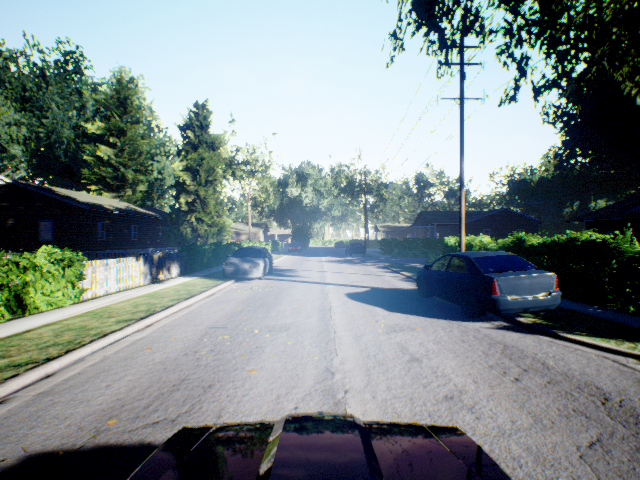}
        \caption{RGB image: Sunny day}
    \end{subfigure}
    \begin{subfigure}{.44\columnwidth}
        \centering
        \includegraphics[width=.95\columnwidth]{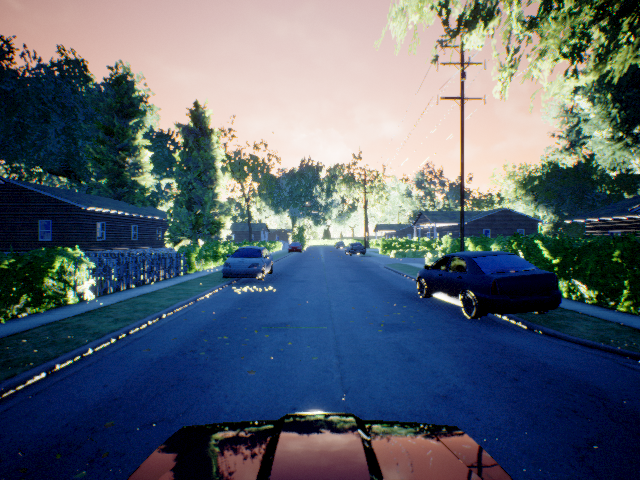}
        \caption{RGB image: Sunrise}
    \end{subfigure}
    \begin{subfigure}{.44\columnwidth}
        \centering
        \includegraphics[width=.95\columnwidth]{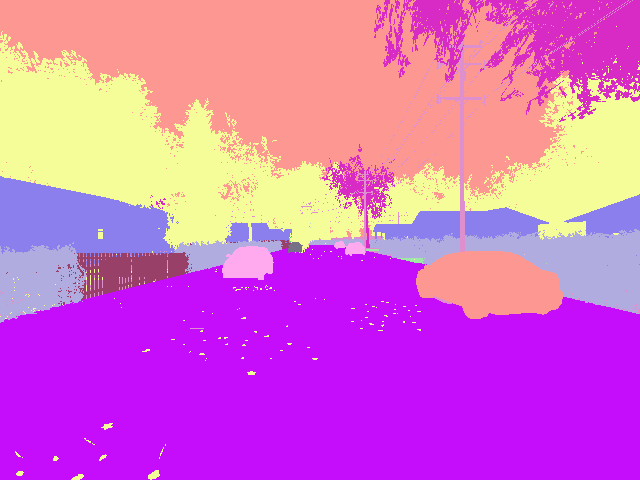}
        \caption{Semantic segmentation}
    \end{subfigure}
    \begin{subfigure}{.44\columnwidth}
        \centering
        \includegraphics[width=.95\columnwidth]{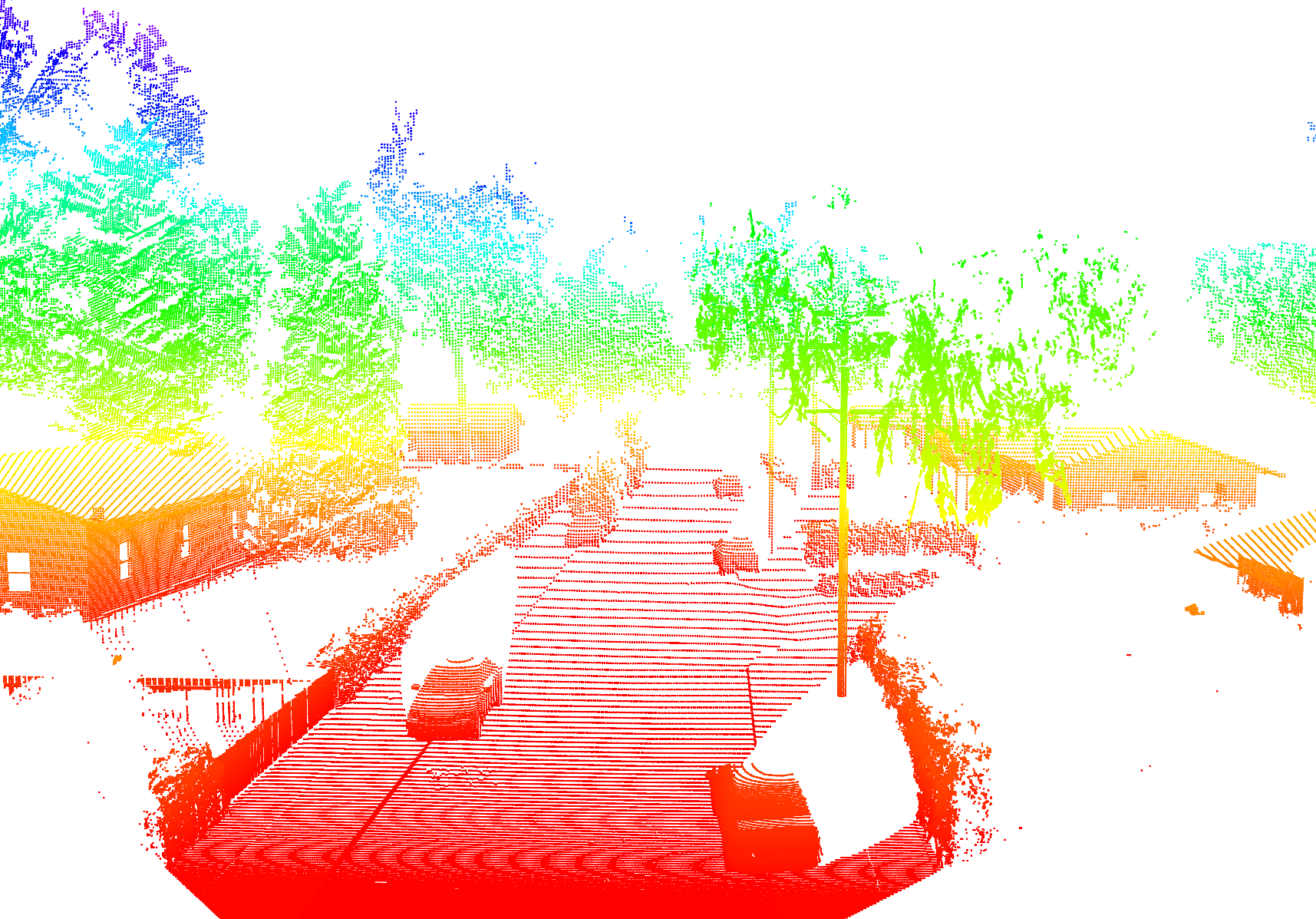}
        \caption{3D \ac{lidar} \ac{pc}}
    \end{subfigure}
    \caption{Overview of simulated sensor modalities and environmental conditions.}
    \label{fig:env-conditions}
    \vspace{-0.3cm}
\end{figure}
An overview of the neighborhood, the simulated environmental conditions, as well as the simulated trajectories, is given in Figure~\ref{fig:neighborhood-overview} and \ref{fig:env-conditions}, and Table~\ref{tab:neighborhood-trajectories}, respectively.

\subsubsection{Real world data}
For demonstrating the applicability of our approach to real-world environments, we use the NCLT dataset \cite{Carlevaris-Bianco2016UniversityDataset}.
The dataset provides raw sensor data from a \ac{lidar} sensor, an omnidirectional camera, and ground truth position data, among others.
As the images lack ground-truth semantic labelling, we used the author's implementation\footnote{\url{https://github.com/NVIDIA/semantic-segmentation}, accessed on 8th of November 2020} from \cite{tao2020hierarchical} to semantically segment the images from the five horizontal color cameras.
The camera-to-\ac{lidar} extrinsics, provided in the dataset, were used to enrich the \acp{pc} with color and semantic information by projection onto the image plane. 
As a last step, we removed the local ground plane from the enriched \ac{pc} in order to make the subsequent segmentation process more robust.
Because the dataset was recorded on a SegWay, which experiences a lot of back and forth pitching motion, the ground plane cannot simply be removed by setting a height threshold on the raw \ac{pc}.
Instead, we estimated the local ground plane based on the points in the enriched \ac{pc} whose class labels correspond to \emph{ground} and subsequently removed all points in close proximity to these points.
A subset of the dataset featuring large map overlap and different environmental conditions was processed and used for our experiments as listed in Table \ref{tab:nclt-trajectories}.

\begin{comment}
\begin{enumerate}
    \item Describe NCLT (environment, dataset, sensors, platform,...)
    \item Define our test sequences (why these, map overlay...)
    \item Processing (semantic segmentation, augmentation, ground segmentation,...)
    \item Peculiarities (Segway, large temporal and spatial scales, qualitative description)
\end{enumerate}
See \href{https://docs.google.com/presentation/d/18yJ8k4K_qnEScfzHHuFX2bqNP3k1e16fF9X8b0uP3jQ/edit?usp=sharing}{here} for raw slides.
\end{comment}

\begin{table}[t]
    \centering
    \vspace{0.3cm}
    \caption{Overview of simulated trajectories. The referenced trajectories as well as environmental conditions are shown in Figures~\ref{fig:neighborhood-overview} and~\ref{fig:env-conditions}, respectively.}
    \begin{tabular}{crll}
    \toprule
    \#  & Length & Conditions & Comment  \\ % bagname
    \midrule
    \texttt{S0} & $\unit[2071]{m}$ & daytime, sunny, clear & Covers whole map\\ % 2020-11-03/full-map.bag
    \texttt{S1} & $\unit[1462]{m}$ & daytime, sunny, clear & Medium size with overlap \\ % 2020-10-28/u-turn.bag
    \texttt{S2} & $\unit[377]{m}$ & daytime, sunny, clear & Single small loop\\ % 2020-10-14/bagfile.bag
    \texttt{S3} & $\unit[358]{m}$ & early sunrise, shadows & Opposite direction to \texttt{S2}\\ % 2020-10-21/sunrise.bag
    \bottomrule
    \end{tabular}
    \label{tab:neighborhood-trajectories}
\end{table}
\begin{table}[t]
    \centering
    \caption{Overview of the trajectories extracted from the NCLT dataset. 
    Please refer to~\cite{Carlevaris-Bianco2016UniversityDataset} for a detailed description of the dataset.}
    \begin{tabular}{ccccl}
    \toprule
    \# & Date & Time & Conditions \\
    \midrule
    \texttt{N0} & 2012-04-29 & $\unit[4]{min}-\unit[30]{min}$ & morning, sunny, foliage \\
    \texttt{N1} & 2012-05-26 &$\unit[0]{min}-\unit[10]{min}$ & evening, sunny, foliage \\
    \texttt{N2} & 2012-11-04 & $\unit[0]{min}-\unit[25]{min}$ & morning, cloudy \\
    \texttt{N3} & 2013-04-05 & $\unit[54]{min}-\unit[69]{min}$ & afternoon, sunny, snow \\
    \bottomrule
    \end{tabular}
    \label{tab:nclt-trajectories}
\end{table}

\subsection{Segmentation}
To evaluate the segmentation method presented in Section~\ref{sec:semseg} we use the simulation dataset run \texttt{S1} and the NCLT run \texttt{N0}.
For the classic Euclidean segmentation in \segmap{} we use the default segmentation distance $d_{segment} = 0.2$.
In \semsegmap{} we adjust this to $d_{segment}=0.3$ and set the penalties and thresholds in the color and semantic distance functions to $t_h=0.1$, $p_h=0.05$ and $p_c=0.15$.

\begin{figure}
    \centering
    \vspace{0.3cm}
    \includegraphics[width=0.9\columnwidth]{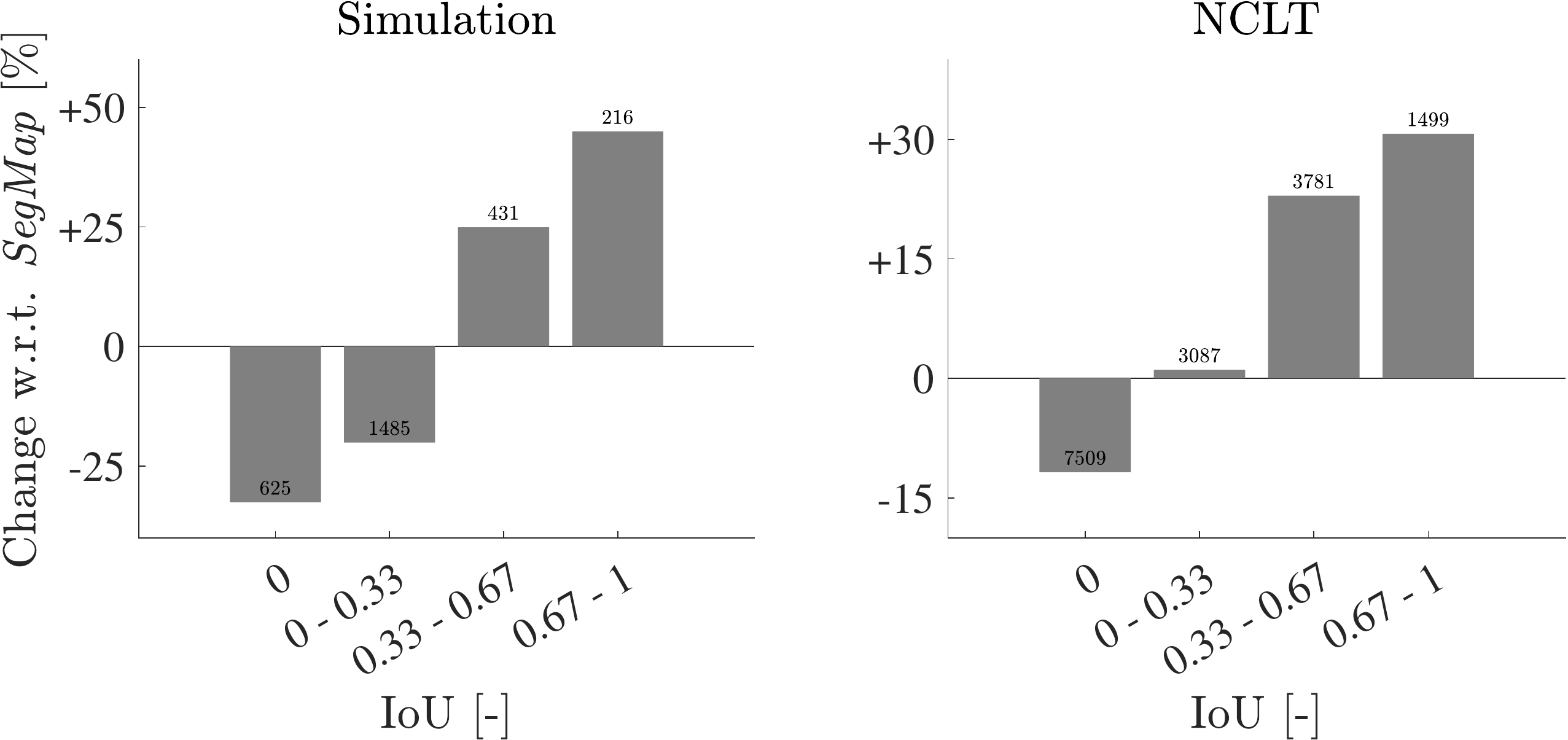}
    \caption{Change of the segmentation \ac{iou} for the simulation dataset \texttt{S1} (left) and NCLT \texttt{N0} (right) using \semsegmap{} with respect to \segmap{}. The bins represent the change in number of segments with a specific \ac{iou} range. The small numbers on top of the bars depict the absolute numbers of segments in that range produced by \semsegmap{}. While \semsegmap{} produces less segments with low \ac{iou}, which represent inconsistent segmentations, it is able to produce more consistent segments with a high overlap and \ac{iou} compared to \segmap{}.}
    \label{fig:iou-segments}
    \vspace{-0.3cm}
\end{figure}

To measure the segmentation quality, we first calculate the convex hull $V_i$ of each segment $S_i$ in the local map, and then compare it to the segment $S_j$ with the convex hull $V_j$ at the same global location in the target map. 
The quality of the segmentation is then given by the \ac{iou} as
\begin{equation}
    \text{IoU}(V_i, V_j) = \frac{V_i \cap V_j}{V_i \cup V_j}.
\end{equation}
In this way, we compare how repeatable and accurate the segmentation is when visiting the same place multiple times.
Figure~\ref{fig:iou-segments} shows the segmentation \ac{iou} results from run S1 in the simulated environment which features multiple intersecting loops in the same environment and N0 for the NCLT dataset.
A low \ac{iou} indicates that the segment was not properly re-segmented when re-visiting the place, while a high \ac{iou} corresponds to a consistent re-segmentation.
In the latter case the segments occupy the same volume in space and are therefore more likely to be matched based on their descriptors.
We consider an $\text{IoU} \geq 0.33$ to represent a good overlap while an $\text{IoU} < 0.33$ represents an inconsistent segmentation.
\semsegmap{} produced $\unit[-24.24]{\%}$ and $\unit[-8.38]{\%}$ less inconsistent segments with $\text{IoU} < 0.33$ while obtaining $\unit[30.97]{\%}$ and $\unit[25.03]{\%}$ more consistent segments with an $\text{IoU} \geq 0.33$ with respect to \segmap{} for the simulation and NCLT dataset, respectively.
\begin{comment}
\begin{figure}
    \centering
    \includegraphics[width=0.95\columnwidth]{images/iou-crop.pdf}
    \caption{Segmentation \ac{iou} for the different datasets. A low \ac{iou} represents inconsistent segments that can not be found again while high \ac{iou} represents segments that get consistently re-detected.}
    \label{fig:iou-segments}
\end{figure}
\end{comment}
%

\subsection{Descriptor quality}
To evaluate the impact of our adapted descriptor extraction network described in Section~\ref{sec:method-description}, we performed a similar experiment as described in~\cite[Section 5.4]{dube2019segmap}. 

\begin{figure}
    \centering
    \vspace{0.3cm}
    \includegraphics[width=0.9\columnwidth]{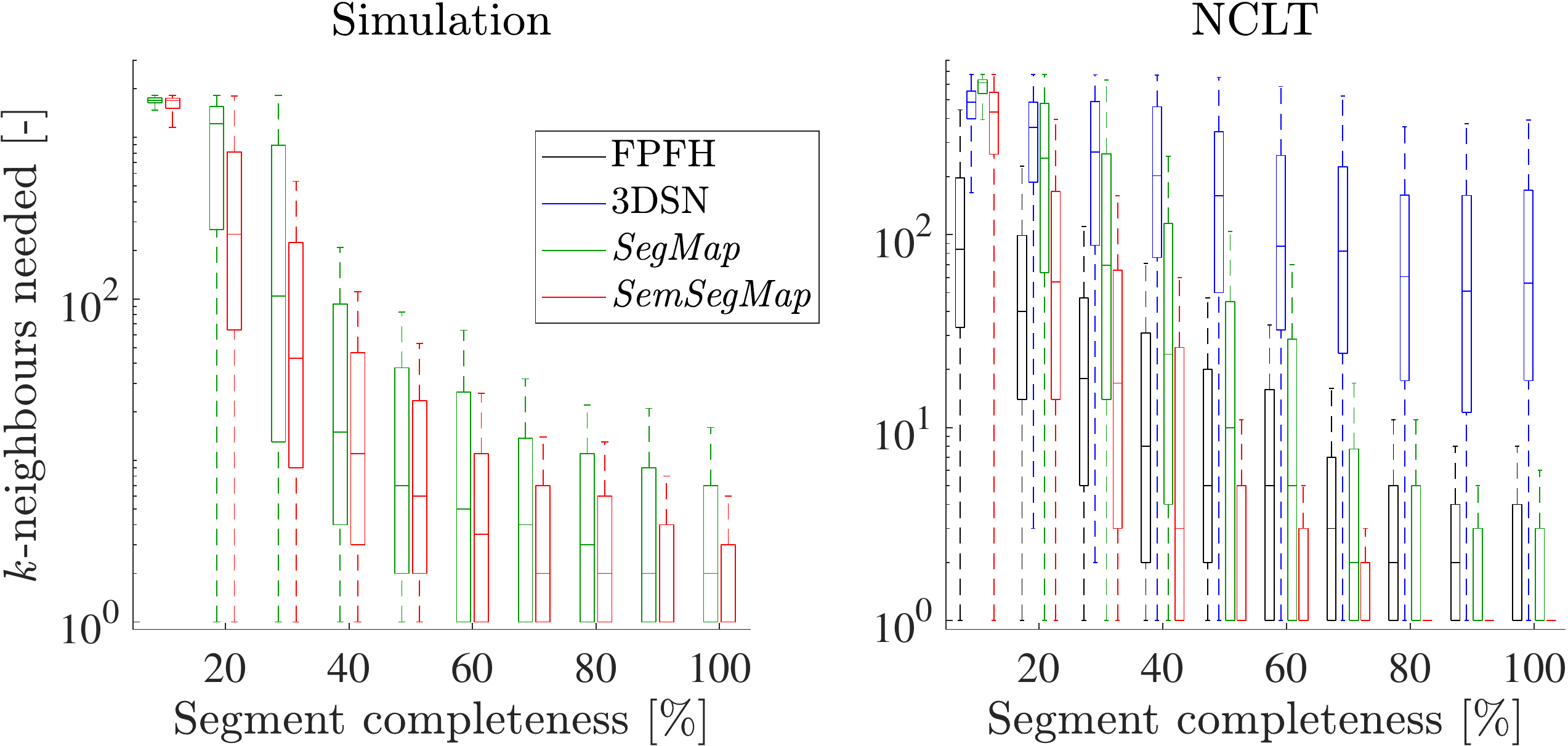}
    \caption{Comparison of descriptor quality on the simulation dataset \texttt{S0} (left) and the NCLT dataset \texttt{N0} (right).}
    \label{fig:descriptor-retrieval}
    \vspace{-0.3cm}
\end{figure}

In order to use the segments as landmarks in a localization and loop-closure scenario as described in Section~\ref{sec:localization}, we obtain candidate matches using \ac{k-nn}.
For a more robust and efficient localization, the amount $k$ of candidates necessary to retrieve the correct match should be as small as possible, even with partially observed segments which occur during live operation or different directions of travel.
In Figure~\ref{fig:descriptor-retrieval}, we show a comparison of how high $k$ needs to be in order to retrieve the correct match using \semsegmap{} and different state-of-the-art \ac{pc} descriptors operating on the simulation dataset \texttt{S0} and the NCLT dataset \texttt{N0}.
The \semsegmap{} network is trained solely on ground truth data obtained from \texttt{S1} and not retrained on the NCLT data.

We compare the quality of our descriptor to a learned segment descriptor~\cite{dube2019segmap} for the simulated data, and extended the evaluation for the NCLT dataset to the hand-crafted descriptor FPFH~\cite{Rusu2009FPFH} and the learned local descriptor 3DSmoothNet~\cite{gojcic2019perfect}.
For FPFH, a single segment descriptor was computed by selecting the segment centroid as the keypoint and choosing the descriptor radius to encompass all points of the respective segment~\cite{anu_nguatem_jan_2015}.
In the case of 3DSmoothNet, simply extending the radius did not yield meaningful results.
One reason could be that the descriptor was specifically designed and trained as a local descriptor.
In order to deploy it within our segment-based framework, we randomly select points from the segment \ac{pc} as keypoints, extract local descriptors for each keypoint and aggregate them to a global descriptor by following a \ac{bow} approach using $k$-means clustering.
Specifically, we trained the $k$-means model on a subset of segments extracted from \texttt{N0}, where $k = 16$ yielded the best results.

Note from Figure~\ref{fig:descriptor-retrieval} that \semsegmap{} outperforms all other descriptors in both datasets except for FPFH with incomplete segments with completeness $<\unit[30]{\%}$. 
This property of the \semsegmap{} descriptor is controlled by the $\sigma$ term in the triplet loss formulation introduced in Equation~\ref{eq:triplet} that biases the network towards better matching more complete segments.
Removing the influence of this term causes better matching performance for harder matches, \ie{} at low completeness thresholds, but at the same time reduces the matching performance of more complete segments as the descriptor attempts to perform more unlikely matches.
However, in a typical \ac{slam} application, while an early localization, \eg{} with many incomplete segments is important, achieving more accurate and robust localizations during operation with complete segments is often of greater interest in a full smoothing and mapping framework.
The descriptor transfers well to the more challenging real-world data from the NCLT dataset, where the performance difference is even more evident.

\subsection{Localization accuracy and robustness}
To assess the localization accuracy and robustness, we tested \semsegmap{s} ability to localize in a previously built target map. 
Therefore, we recorded the target map on trajectories \texttt{S0} and \texttt{N0} for the simulation and NCLT dataset, respectively.
The map for \texttt{S0} contains $2006$ segments and has a size of $\unit[0.51]{MB}$ while for \texttt{N0} there are $2588$ segments in a $\unit[0.66]{MB}$ map.
All other trajectories except for the training set \texttt{S1} are used for on-line localization against the target map as described in Section~\ref{sec:localization}. 

For the localization evaluation we retrieve a total of 16 neighbours using \ac{k-nn}.
For the geometric verification in our experiments the \ac{ransac} was set to require a minimum of $6$ and $7$ inliers for the simulation and NCLT dataset, respectively, and allow a centroid distance of at most $\unit[0.4]{m}$.
For all the experiments for both \segmap{} and \semsegmap{} we keep these parameters fixed and only change the segmentation and description process as outlined in Sections~\ref{sec:method-segmentation} and \ref{sec:method-description}, to provide the fairest comparison.
With these settings \semsegmap{} (excluding the semantic segmentation) runs on an Intel i7-8700 CPU with an average frequency of $\unit[6.13]{Hz}$ and $\unit[6.74]{Hz}$ on the simulation and NCLT dataset, respectively.

\begin{figure}
    \centering
    \vspace{0.3cm}
    \includegraphics[width=0.9\columnwidth]{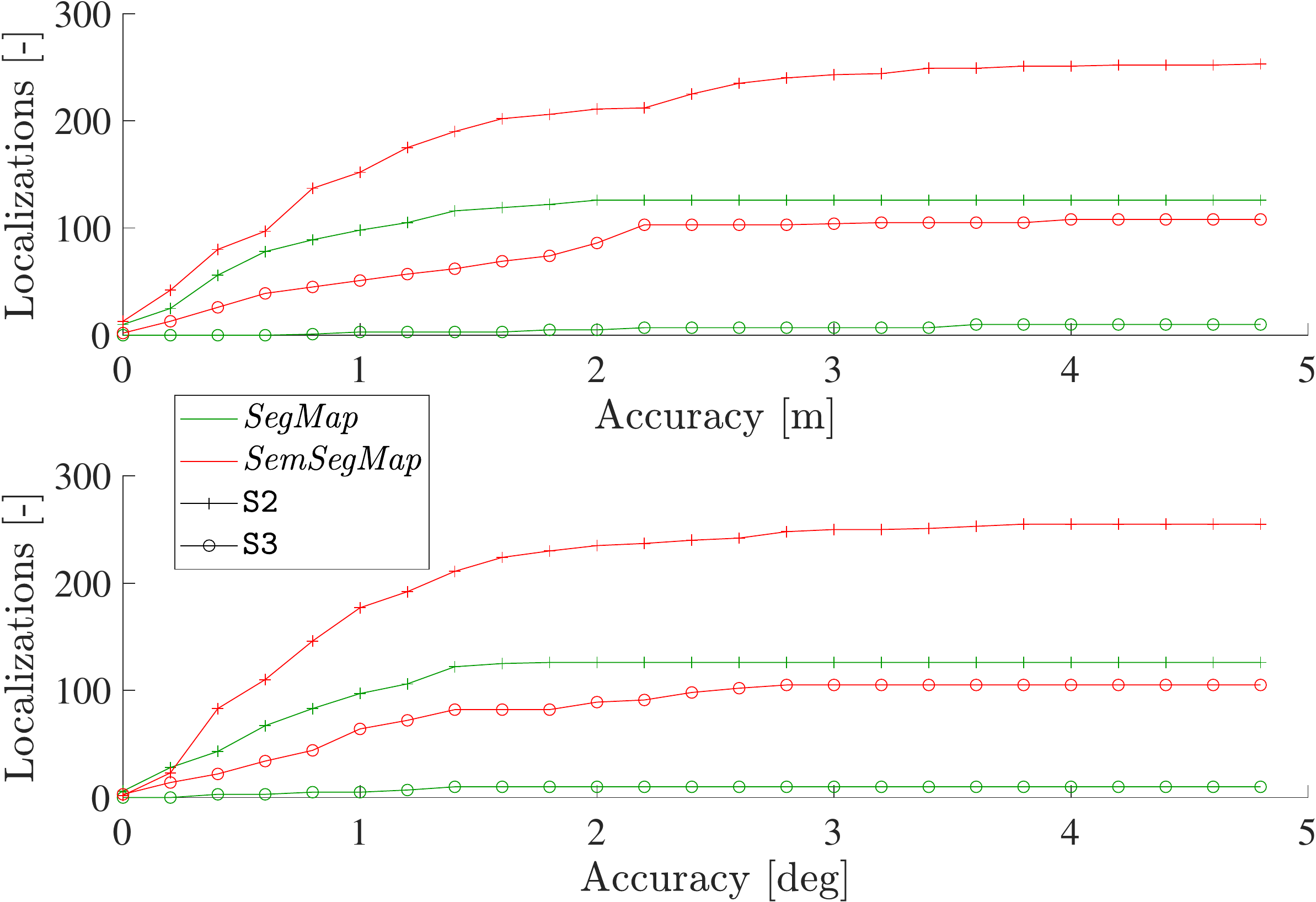}
    \caption{Cumulative successful localizations with a certain accuracy on the simulation datasets with a target map built from \texttt{S0}.}
    \label{fig:acc-recall-simulation}
\end{figure}

\begin{figure}
    \centering
    \includegraphics[width=0.9\columnwidth]{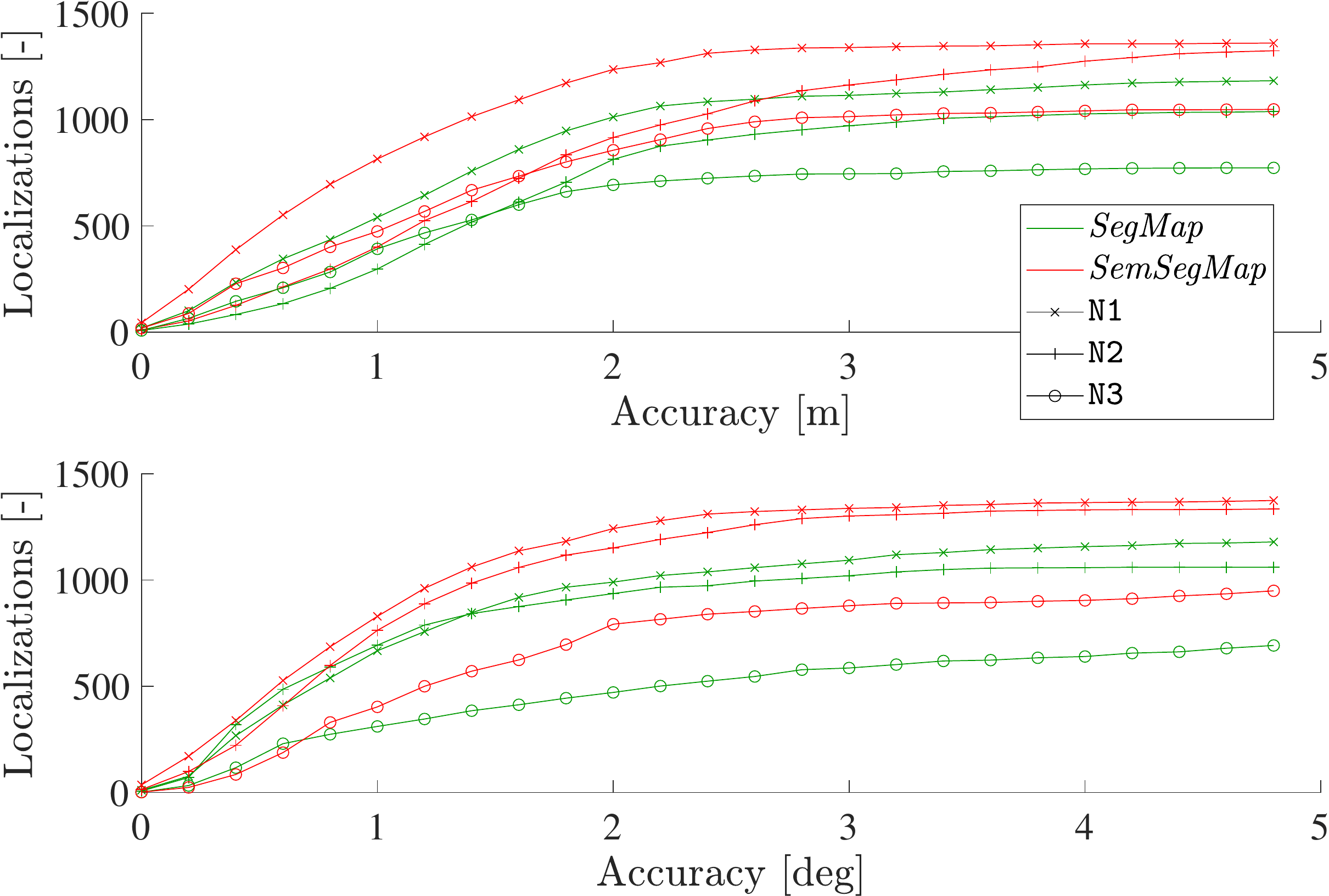}
    \caption{Cumulative successful localizations with a certain accuracy on the real-world datasets with a target map built from \texttt{N0}.}
    \label{fig:acc-recall-nclt}
    \vspace{-0.3cm}
\end{figure}

Table~\ref{tab:localization-accuracy} and Figures~\ref{fig:acc-recall-simulation} and~\ref{fig:acc-recall-nclt} report the accuracy results of the estimated 6-\ac{dof} pose for the simulation and NCLT dataset, respectively.
\semsegmap{} is able to consistently find more localizations throughout all the tested datasets.
In simulation, less affected by odometry and sensor noise, \semsegmap{} is able to find a total of $\unit[102]{\%}$ more high accuracy localizations (translation error of less than $\unit[1]{m}$) with respect to \segmap{} and $\unit[165]{\%}$ more accurate localizations (translation error of less than $\unit[5]{m}$).
Especially on trajectory \texttt{S3}, \segmap{} suffers from different appearance and viewpoint while \semsegmap{} is less affected and still able to produce many accurate localizations.
Those results also transfer to the real-world dataset where more than $\unit[50.9]{\%}$ high accuracy localization and $\unit[24.7]{\%}$ more accurate localizations are found compared to \segmap.
\begin{table}
    \centering
    \vspace{0.3cm}
    \caption{Localization accuracy results overview ($n_{\unit[<\bullet]{m}}$ relative improvement of \semsegmap{} with respect to \segmap{} with a certain accuracy).}
    \begin{tabular}{lrrrr}
    \toprule
    \#  & \multicolumn{2}{c}{Number of localizations} &  $n_{\unit[<1]{m}}$ &$n_{\unit[<5]{m}}$  \\%& \multicolumn{2}{c}{Avg. inliers}  \\ 
    &\segmap&\semsegmap&&\\%&SM&SSM\\
    \midrule
    \texttt{S2}  & 126 & 258 & \unit[53.93]{\%} & \unit[100.79]{\%}  \\%& 7.38 & 8.16\\
    \texttt{S3}  & 10 & 108 & \unit[4400]{\%} & \unit[980]{\%} \\%& 6.20 & 6.42 \\
    \midrule
    \texttt{N1}  & 1201 & 1395 & \unit[60.00]{\%} & \unit[14.96]{\%} \\%& 9.27 & 10.99 \\
    \texttt{N2}  & 1060 & 1337 & \unit[44.17]{\%} & \unit[27.55]{\%} \\%& 9.37 & 10.27 \\
    \texttt{N3}  & 774 & 1056 & \unit[41.70]{\%} & \unit[35.58]{\%} \\%& 9.26 & 9.85 \\
    \bottomrule
    \end{tabular}
    \label{tab:localization-accuracy}
    \vspace{-0.3cm}
\end{table}

\begin{comment}
\begin{table}
    \centering
    \caption{Localization accuracy results overview ($n_{loc}$: Number of localizations, $\bar{e}_t$: Average translation error, $\bar{e}_r$: Average rotation error, SM: \segmap, SSM: \semsegmap).}
    \begin{tabular}{lrrrrrr}
    \toprule
    \#  & \multicolumn{2}{c}{$n_{loc}$} & \multicolumn{2}{c}{$\bar{e}_t [\unit{m}]$} & \multicolumn{2}{c}{$\bar{e}_r [\unit{deg}]$} \\%& \multicolumn{2}{c}{Avg. inliers}  \\ 
    &SM&SSM&SM&SSM&SM&SSM%\\&SM&SSM\\
    \midrule
    \texttt{S2}  & 126 & 258 & 0.80 & 1.36 & 0.81 & 1.14 \\%& 7.38 & 8.16\\
    \texttt{S3}  & 10 & 108 & 2.25 & 1.41 & 1.06 & 1.36 \\%& 6.20 & 6.42 \\
    \midrule
    \texttt{N1}  & 1201 & 1395 & 1.48 & 1.26 & 1.44 & 1.26 \\%& 9.27 & 10.99 \\
    \texttt{N2}  & 1060 & 1337 & 1.78 & 1.90 & 1.14 & 1.29 \\%& 9.37 & 10.27 \\
    \texttt{N3}  & 774 & 1056 & 1.33 & 1.45 & 2.24 & 2.06 \\%& 9.26 & 9.85 \\
    \bottomrule
    \end{tabular}
    \label{tab:localization-accuracy}
\end{table}
\end{comment}

\section{Conclusions} \label{sec:conclusions}
% Everyone at the end
In this paper, we introduced \semsegmap{}, an extension to \segmap{} that fuses both color and semantic information from an RGB camera with \ac{lidar} data in real-time.
In a real-world robotic application, the addition of cameras to a platform equipped with a \ac{lidar} is typically easily possible due to the comparably low price of cameras and their cross-purpose use, especially when also performing semantic segmentation to improve scene understanding.
We include this additional modality both to improve segmentation and descriptor quality, which we showed in a stimulated dataset with accurate ground truth and a challenging real-world dataset.

Using the described extensions, \semsegmap{} is able to outperform a geometric segmentation approach by producing less inconsistent segments and more highly overlapping segments when re-visiting a place.
The tight fusion of the additional information in the descriptor also increases descriptor quality where \semsegmap{} not only outperforms the \segmap{} baseline but also other state-of-the-art \ac{pc} descriptors like FPFH and 3DSmoothNet in terms of \ac{k-nn} required to find the correct match.
These improvements also propagate to the localization accuracy and robustness resulting in \semsegmap{} providing $\unit[102]{\%}$ and $\unit[50.9]{\%}$ more high accuracy localizations than \segmap{} for the simulated and the real-world dataset, respectively.

To further extend our framework, a combination of FPFH and \semsegmap{} descriptor based on the expected completeness of a segment could be used in order to benefit from both advantages.


\begin{acronym}
    \acro{gnss}[GNSS]{Global Navigation Satellite System}
    \acro{imu}[IMU]{Inertial Measurement Unit}
    \acro{slam}[SLAM]{Simultaneous Localization and Mapping}
    \acro{mav}[MAV]{Micro Aerial Vehicle}
    \acro{fov}[FoV]{Field of View}
    \acro{rtk}[RTK]{Real-Time Kinematic}
    \acro{rmse}[RMSE]{Root Mean Square Error}
    \acro{lidar}[LiDAR]{Light Detection and Ranging}
    \acro{icp}[ICP]{Iterative Closest Point}
    \acro{k-nn}[$k$-NN]{$k$ nearest neighbours}
    \acro{dof}[DoF]{degree of freedom}
    \acro{ransac}[RANSAC]{Random Sample Consensus}
    \acro{iou}[IoU]{Intersection Over Union}
    \acro{pc}[PC]{point cloud}
    \acro{bow}[BoW]{bag of words}
\end{acronym}
%%%%%%%%%%%%%%%%%%%%%%%%%%%%%%%%%%%%%%%%%%%%%%%%%%%%%%%%%%%%%%%%%%%%%%%%%%%%%%%
% Bibliography
%%%%%%%%%%%%%%%%%%%%%%%%%%%%%%%%%%%%%%%%%%%%%%%%%%%%%%%%%%%%%%%%%%%%%%%%%%%%%%%
\bibliographystyle{IEEEtran}
\bibliography{references.bib}

\begin{thebibliography}{10}
\providecommand{\url}[1]{#1}
\csname url@samestyle\endcsname
\providecommand{\newblock}{\relax}
\providecommand{\bibinfo}[2]{#2}
\providecommand{\BIBentrySTDinterwordspacing}{\spaceskip=0pt\relax}
\providecommand{\BIBentryALTinterwordstretchfactor}{4}
\providecommand{\BIBentryALTinterwordspacing}{\spaceskip=\fontdimen2\font plus
\BIBentryALTinterwordstretchfactor\fontdimen3\font minus
  \fontdimen4\font\relax}
\providecommand{\BIBforeignlanguage}[2]{{%
\expandafter\ifx\csname l@#1\endcsname\relax
\typeout{** WARNING: IEEEtran.bst: No hyphenation pattern has been}%
\typeout{** loaded for the language `#1'. Using the pattern for}%
\typeout{** the default language instead.}%
\else
\language=\csname l@#1\endcsname
\fi
#2}}
\providecommand{\BIBdecl}{\relax}
\BIBdecl

\bibitem{Cadena2016}
C.~Cadena, L.~Carlone, H.~Carrillo, Y.~Latif, D.~Scaramuzza, J.~Neira, I.~Reid,
  and J.~J. Leonard, ``{Past, Present, and Future of Simultaneous Localization
  and Mapping: Toward the Robust-Perception Age},'' \emph{IEEE Transactions on
  Robotics}, vol.~32, no.~6, Dec. 2016.

\bibitem{Schneider2017}
T.~Schneider, M.~Dymczyk, M.~Fehr, K.~Egger, S.~Lynen, I.~Gilitschenski, and
  R.~Siegwart, ``{maplab: An Open Framework for Research in Visual-inertial
  Mapping and Localization},'' \emph{IEEE Robotics and Automation Letters},
  vol.~3, no.~3, 11 2018.

\bibitem{Mur-Artal2015}
R.~Mur-Artal, J.~M.~M. Montiel, and J.~D. Tardos, ``{ORB-SLAM: A Versatile and
  Accurate Monocular SLAM System},'' \emph{IEEE Transactions on Robotics},
  vol.~31, no.~5, 10 2015.

\bibitem{Milford2012}
M.~J. Milford and G.~F. Wyeth, ``{SeqSLAM: Visual route-based navigation for
  sunny summer days and stormy winter nights},'' in \emph{IEEE International
  Conference on Robotics and Automation (ICRA)}, 2012.

\bibitem{RevaudR2D2:Descriptor}
J.~Revaud, P.~Weinzaepfel, C.~R. de~Souza, and M.~Humenberger, ``{R2D2:}
  repeatable and reliable detector and descriptor,'' in \emph{Advances in
  Neural Information Processing Systems}, 2019.

\bibitem{Sarlin2018FromScale}
P.-E. Sarlin, C.~Cadena, R.~Siegwart, and M.~Dymczyk, ``{From Coarse to Fine:
  Robust Hierarchical Localization at Large Scale},'' in \emph{IEEE Computer
  Society Conference on Computer Vision and Pattern Recognition}, Dec. 2018.

\bibitem{Arandjelovic2015}
R.~{Arandjelović}, P.~{Gronat}, A.~{Torii}, T.~{Pajdla}, and J.~{Sivic},
  ``{NetVLAD: CNN Architecture for Weakly Supervised Place Recognition},''
  \emph{IEEE Transactions on Pattern Analysis and Machine Intelligence},
  vol.~40, no.~6, 2018.

\bibitem{dube2019segmap}
R.~Dub{\'e}, A.~Cramariuc, D.~Dugas, H.~Sommer, M.~Dymczyk, J.~Nieto,
  R.~Siegwart, and C.~Cadena, ``{SegMap: Segment-based mapping and localization
  using data-driven descriptors},'' \emph{The International Journal of Robotics
  Research}, vol.~39, no. 2-3, 2019.

\bibitem{Elhousni2020AVehicles}
M.~Elhousni and X.~Huang, ``{A Survey on 3D LiDAR Localization for Autonomous
  Vehicles},'' in \emph{2020 IEEE Intelligent Vehicles Symposium (IV2020)}, Las
  Vegas, US, May 2020.

\bibitem{noh_large-scale_2017}
H.~Noh, A.~Araujo, J.~Sim, T.~Weyand, and B.~Han, ``Large-{Scale} {Image}
  {Retrieval} with {Attentive} {Deep} {Local} {Features},'' in \emph{2017
  {IEEE} {International} {Conference} on {Computer} {Vision} ({ICCV})}, Oct.
  2017.

\bibitem{Gawel2017}
A.~Gawel, C.~Del~Don, R.~Siegwart, J.~Nieto, and C.~Cadena, ``{X-View:
  Graph-Based Semantic Multi-View Localization},'' \emph{IEEE Robotics and
  Automation Letters}, vol.~3, no.~3, 2017.

\bibitem{benbihi_image-based_2020}
A.~Benbihi, S.~Arravechia, M.~Geist, and C.~Pradalier, ``Image-{Based} {Place}
  {Recognition} on {Bucolic} {Environment} {Across} {Seasons} {From} {Semantic}
  {Edge} {Description},'' in \emph{2020 IEEE International Conference on
  Robotics and Automation (ICRA)}, 2020.

\bibitem{hu2020dasgil}
H.~Hu, Z.~Qiao, M.~Cheng, Z.~Liu, and H.~Wang, ``Dasgil: Domain adaptation for
  semantic and geometric-aware image-based localization,'' \emph{IEEE
  Transactions on Image Processing}, vol.~30, 2020.

\bibitem{lowry_visual_2016}
S.~Lowry, N.~Sunderhauf, P.~Newman, J.~J. Leonard, D.~Cox, P.~Corke, and M.~J.
  Milford, ``Visual {Place} {Recognition}: {A} {Survey},'' \emph{IEEE
  Transactions on Robotics}, vol.~32, no.~1, Feb. 2016.

\bibitem{noauthor_benchmarking_nodate}
L.~Hammarstrand, F.~Kahl, W.~Maddern, T.~Pajdla, M.~Pollefeys, T.~Sattler,
  J.~Sivic, E.~Stenborg, C.~Toft, and A.~Torii, ``Benchmarking {Long}-term
  {Visual} {Localization},'' \url{https://www.visuallocalization.net/}.

\bibitem{Ok2019RobustNavigation}
K.~Ok, K.~Liu, K.~Frey, J.~P. How, and N.~Roy, ``{Robust Object-based SLAM for
  High-speed Autonomous Navigation},'' in \emph{International Conference on
  Robotics and Automation}, Montreal, Canada, May 2019.

\bibitem{taubner_lcd_2020}
F.~Taubner, F.~Tschopp, T.~Novkovic, R.~Siegwart, and F.~Furrer, ``{LCD} --
  {Line} {Clustering} and {Description} for {Place} {Recognition},'' in
  \emph{International Virtual Conference on 3D Vision (3DV)}, Fukuoka, Japan,
  Nov. 2020.

\bibitem{besl1992method}
P.~J. Besl and N.~D. McKay, ``Method for registration of 3-d shapes,'' in
  \emph{Sensor fusion IV: control paradigms and data structures}, vol. 1611,
  1992.

\bibitem{bernreiter2021phaser}
L.~Bernreiter, L.~Ott, J.~Nieto, R.~Siegwart, and C.~Cadena, ``{PHASER: a
  Robust and Correspondence-free Global Pointcloud Registration},'' \emph{IEEE
  Robotics and Automation Letters}, vol.~6, no.~2, 2021.

\bibitem{Le_2019_CVPR}
H.~M. Le, T.-T. Do, T.~Hoang, and N.-M. Cheung, ``{SDRSAC: Semidefinite-Based
  Randomized Approach for Robust Point Cloud Registration Without
  Correspondences},'' in \emph{IEEE/CVF Conference on Computer Vision and
  Pattern Recognition (CVPR)}, June 2019.

\bibitem{chen_overlapnet_2020}
X.~Chen, T.~Läbe, A.~Milioto, T.~Röhling, O.~Vysotska, A.~Haag, J.~Behley,
  and C.~Stachniss, ``\BIBforeignlanguage{en}{{OverlapNet}: {Loop} {Closing}
  for {LiDAR}-based {SLAM}},'' in \emph{\BIBforeignlanguage{en}{Robotics:
  {Science} and {Systems} {XVI}}}, Jul. 2020.

\bibitem{yin20193d}
H.~Yin, Y.~Wang, X.~Ding, L.~Tang, S.~Huang, and R.~Xiong, ``{3D LiDAR-Based
  Global Localization Using Siamese Neural Network},'' \emph{IEEE Transactions
  on Intelligent Transportation Systems}, vol.~21, no.~4, 2019.

\bibitem{kim_1-day_2019}
G.~Kim, B.~Park, and A.~Kim, ``1-{Day} {Learning}, 1-{Year} {Localization}:
  {Long}-{Term} {LiDAR} {Localization} {Using} {Scan} {Context} {Image},''
  \emph{IEEE Robotics and Automation Letters}, vol.~4, no.~2, Apr. 2019.

\bibitem{Yin2018LocNet}
H.~{Yin}, L.~{Tang}, X.~{Ding}, Y.~{Wang}, and R.~{Xiong}, ``Locnet: Global
  localization in 3d point clouds for mobile vehicles,'' in \emph{2018 IEEE
  Intelligent Vehicles Symposium (IV)}, 2018, pp. 728--733.

\bibitem{Rusu2009FPFH}
R.~B. {Rusu}, N.~{Blodow}, and M.~{Beetz}, ``{Fast Point Feature Histograms
  (FPFH) for 3D registration},'' in \emph{2009 IEEE International Conference on
  Robotics and Automation}, 2009.

\bibitem{lu2019l3}
W.~Lu, Y.~Zhou, G.~Wan, S.~Hou, and S.~Song, ``L3-net: Towards learning based
  lidar localization for autonomous driving,'' in \emph{IEEE/CVF Conference on
  Computer Vision and Pattern Recognition}, 2019.

\bibitem{kallasi2016fast}
F.~Kallasi, D.~L. Rizzini, and S.~Caselli, ``Fast keypoint features from laser
  scanner for robot localization and mapping,'' \emph{IEEE Robotics and
  Automation Letters}, 2016.

\bibitem{gojcic2019perfect}
Z.~Gojcic, C.~Zhou, J.~D. Wegner, and A.~Wieser, ``The perfect match: 3d point
  cloud matching with smoothed densities,'' in \emph{IEEE/CVF Conference on
  Computer Vision and Pattern Recognition}, 2019.

\bibitem{schaupp_oreos_2019}
L.~Schaupp, M.~Bürki, R.~Dubé, R.~Siegwart, and C.~Cadena, ``{OREOS}:
  {Oriented} {Recognition} of {3D} {Point} {Clouds} in {Outdoor} {Scenarios},''
  in \emph{{IEEE}/{RSJ} {International} {Conference} on {Intelligent} {Robots}
  and {Systems} ({IROS})}, Nov. 2019.

\bibitem{zaganidis_semantically_2019}
A.~Zaganidis, A.~Zerntev, T.~Duckett, and G.~Cielniak, ``Semantically
  {Assisted} {Loop} {Closure} in {SLAM} {Using} {NDT} {Histograms},'' in
  \emph{2019 {IEEE}/{RSJ} {International} {Conference} on {Intelligent}
  {Robots} and {Systems} ({IROS})}, Nov. 2019.

\bibitem{douillard2012scan}
B.~Douillard, A.~Quadros, P.~Morton, J.~P. Underwood, M.~De~Deuge, S.~Hugosson,
  M.~Hallstr{\"o}m, and T.~Bailey, ``{Scan segments matching for pairwise 3D
  alignment},'' in \emph{IEEE International Conference on Robotics and
  Automation}, 2012.

\bibitem{nieto2006scan}
J.~Nieto, T.~Bailey, and E.~Nebot, ``{Scan-SLAM: Combining EKF-SLAM and scan
  correlation},'' in \emph{Field and service robotics}, 2006.

\bibitem{tinchev_seeing_2018}
G.~Tinchev, S.~Nobili, and M.~Fallon, ``Seeing the {Wood} for the {Trees}:
  {Reliable} {Localization} in {Urban} and {Natural} {Environments},'' in
  \emph{{IEEE}/{RSJ} {International} {Conference} on {Intelligent} {Robots} and
  {Systems} ({IROS})}, Oct. 2018.

\bibitem{tinchev_learning_2019}
G.~Tinchev, A.~Penate-Sanchez, and M.~Fallon, ``Learning to {See} the {Wood}
  for the {Trees}: {Deep} {Laser} {Localization} in {Urban} and {Natural}
  {Environments} on a {CPU},'' \emph{IEEE Robotics and Automation Letters},
  vol.~4, no.~2, Apr. 2019.

\bibitem{sun_recurrent-octomap_2018}
L.~Sun, Z.~Yan, A.~Zaganidis, C.~Zhao, and T.~Duckett, ``Recurrent-{OctoMap}:
  {Learning} {State}-{Based} {Map} {Refinement} for {Long}-{Term} {Semantic}
  {Mapping} {With} 3-{D}-{Lidar} {Data},'' \emph{IEEE Robotics and Automation
  Letters}, vol.~3, no.~4, Oct. 2018.

\bibitem{zaganidis_integrating_2018}
A.~Zaganidis, L.~Sun, T.~Duckett, and G.~Cielniak, ``Integrating {Deep}
  {Semantic} {Segmentation} {Into} 3-{D} {Point} {Cloud} {Registration},''
  \emph{IEEE Robotics and Automation Letters}, vol.~3, no.~4, Oct. 2018.

\bibitem{chen_suma_2019}
X.~Chen, A.~Milioto, E.~Palazzolo, P.~Giguère, J.~Behley, and C.~Stachniss,
  ``{SuMa}++: {Efficient} {LiDAR}-based {Semantic} {SLAM},'' in
  \emph{{IEEE}/{RSJ} {International} {Conference} on {Intelligent} {Robots} and
  {Systems} ({IROS})}, Nov. 2019.

\bibitem{parkison_semantic_2018}
S.~A. Parkison, L.~Gan, M.~G. Jadidi, and R.~Eustice, ``Semantic {Iterative}
  {Closest} {Point} through {Expectation}-{Maximization},'' in \emph{{BMVC}},
  2018.

\bibitem{Bernreiter2021S2Loc}
L.~Bernreiter, L.~Ott, J.~Nieto, R.~Siegwart, and C.~Cadena, ``{Spherical
  Multi-Modal Place Recognition for Heterogeneous Sensor Systems},'' in
  \emph{International Conference on Robotics and Automation (ICRA)}, 2021.

\bibitem{Ratz2020Oneshot}
S.~Ratz, M.~Dymczyk, R.~Siegwart, and R.~Dub{\'e}, ``{OneShot Global
  Localization: Instant LiDAR-Visual Pose Estimation},'' in \emph{IEEE
  International Conference on Robotics and Automation (ICRA)}, 2020.

\bibitem{Schonberger2017}
J.~L. {Schönberger}, M.~{Pollefeys}, A.~{Geiger}, and T.~{Sattler},
  ``{Semantic Visual Localization},'' in \emph{2018 IEEE/CVF Conference on
  Computer Vision and Pattern Recognition}, 2018.

\bibitem{Qi2017Pointnetpp}
C.~R. Qi, L.~Yi, H.~Su, and L.~J. Guibas, ``Pointnet++: Deep hierarchical
  feature learning on point sets in a metric space,'' in \emph{Advances in
  Neural Information Processing Systems}, 2017.

\bibitem{Kaess2012iSAM2}
M.~Kaess, H.~Johannsson, R.~Roberts, V.~Ila, J.~J. Leonard, and F.~Dellaert,
  ``{iSAM2: Incremental smoothing and mapping using the Bayes tree},''
  \emph{The International Journal of Robotics Research}, vol.~31, no.~2, 2012.

\bibitem{Koenig2004gazebo}
N.~Koenig and A.~Howard, ``Design and use paradigms for gazebo, an open-source
  multi-robot simulator,'' in \emph{IEEE/RSJ International Conference on
  Intelligent Robots and Systems}, Sendai, Japan, Sep 2004.

\bibitem{Dosovitskiy17carla}
A.~Dosovitskiy, G.~Ros, F.~Codevilla, A.~Lopez, and V.~Koltun, ``{CARLA}: {An}
  open urban driving simulator,'' in \emph{1st Annual Conference on Robot
  Learning}, 2017.

\bibitem{Rong2020lgsvl}
G.~{Rong}, B.~H. {Shin}, H.~{Tabatabaee}, Q.~{Lu}, S.~{Lemke}, M.~{Možeiko},
  E.~{Boise}, G.~{Uhm}, M.~{Gerow}, S.~{Mehta}, E.~{Agafonov}, T.~H. {Kim},
  E.~{Sterner}, K.~{Ushiroda}, M.~{Reyes}, D.~{Zelenkovsky}, and S.~{Kim},
  ``{LGSVL} simulator: A high fidelity simulator for autonomous driving,'' in
  \emph{2020 IEEE 23rd International Conference on Intelligent Transportation
  Systems (ITSC)}, 2020.

\bibitem{Shah2017airsim}
S.~Shah, D.~Dey, C.~Lovett, and A.~Kapoor, ``Airsim: High-fidelity visual and
  physical simulation for autonomous vehicles,'' in \emph{Field and Service
  Robotics}, 2017.

\bibitem{Carlevaris-Bianco2016UniversityDataset}
N.~Carlevaris-Bianco, A.~K. Ushani, and R.~M. Eustice, ``{University of
  Michigan North Campus long-term vision and lidar dataset},''
  \emph{International Journal of Robotics Research}, Aug. 2016.

\bibitem{tao2020hierarchical}
A.~Tao, K.~Sapra, and B.~Catanzaro, ``Hierarchical multi-scale attention for
  semantic segmentation,'' 2020, arxiv: 2005.10821.

\bibitem{anu_nguatem_jan_2015}
\BIBentryALTinterwordspacing
Anu, W.~Nguatem, and Jan, ``Computing a global descriptor with local
  descriptors,'' May 2015. [Online]. Available:
  \url{http://www.pcl-users.org/Computing-a-global-descriptor-with-local-descriptors-td4038260.html}
\BIBentrySTDinterwordspacing

\end{thebibliography}
\end{document}